\documentclass{svproc}
\usepackage{url}

\usepackage{subfigure}
\usepackage{hhline}
\usepackage{caption}
\usepackage{graphicx}
\usepackage{amsmath}
\usepackage{multirow}
\usepackage{cite}

\begin{document}
\mainmatter              
\title{Incorporating Orientations into End-to-end Driving Model for Steering Control}
\titlerunning{End-to-end Driving Model}  
%
\author{Peng Wan \and Zhenbo Song \and Jianfeng Lu}

%

\institute{
    \email{{pengw, songzb, lujf}@njust.edu.cn} \\ 
}
%
%

\maketitle              

\begin{abstract}
In this paper, we present a novel end-to-end deep neural network model for autonomous driving that takes monocular image sequence as input, and directly generates the steering control angle. Firstly, we model the end-to-end driving problem as a local path planning process. Inspired by the environmental representation in the classical planning algorithms(\textit{i}.\textit{e}. the beam curvature method), pixel-wise orientations are fed into the network to learn direction-aware features. Next, to handle with the imbalanced distribution of steering values in training datasets, we propose an improvement on a cost-sensitive loss function named SteeringLoss2. Besides, we also present a new end-to-end driving dataset, which provides corresponding LiDAR and image sequences, as well as standard driving behaviors. Our dataset includes multiple driving scenarios, such as urban, country, and off-road. Numerous experiments are conducted on both public available LiVi-Set and our own dataset, and the results show that the model using our proposed methods can predict steering angle accurately.
\end{abstract}
\section{Introduction}
At present, autonomous driving frameworks can be divided into two technical directions: mediated approaches and end-to-end driving models. Mediated approaches first perceive and model the surrounding environment based on the information from multiple sensors, such as camera, LiDAR, and radar. Then path planning algorithms are conducted based on the environmental model, and then the controlling signals including steering angle and speed value are consequently generated. Mediated approaches are often more controllable and robust, because the environmental model has more spatial and semantic information(\textit{i}.\textit{e}. road boundary). Compared with the mediated approaches, with the effectiveness of popular deep neural networks in visual perception tasks, end-to-end models can utilize the raw data from sensors and directly output the controlling signals. In this paper, we focus on the deep end-to-end driving task, which only depends on the monocular video and predicts the steering angle specifically.

For monocular video-based deep end-to-end driving, Nvidia proposed the PilotNet~\cite{bojarski2016end} to estimate the steering angle on the current frame. However, single-frame prediction lacks temporal information, which is not consistent with humans' perception and driving behaviors. Hence, new neural networks(\textit{i}.\textit{e}. FCN-LSTM~\cite{xu2017end}, DBNet~\cite{chen2018lidar}) are designed upon LSTM \cite{hochreiter1997long} models to capture both visual cues and temporal dependencies. Moreover, Fernando~\cite{fernando2017going} incorporated the previous steering wheel trajectory into LSTM to further stress the consistency of controlling signals.

\begin{figure}
	\centering
	\includegraphics[width=0.9\linewidth]{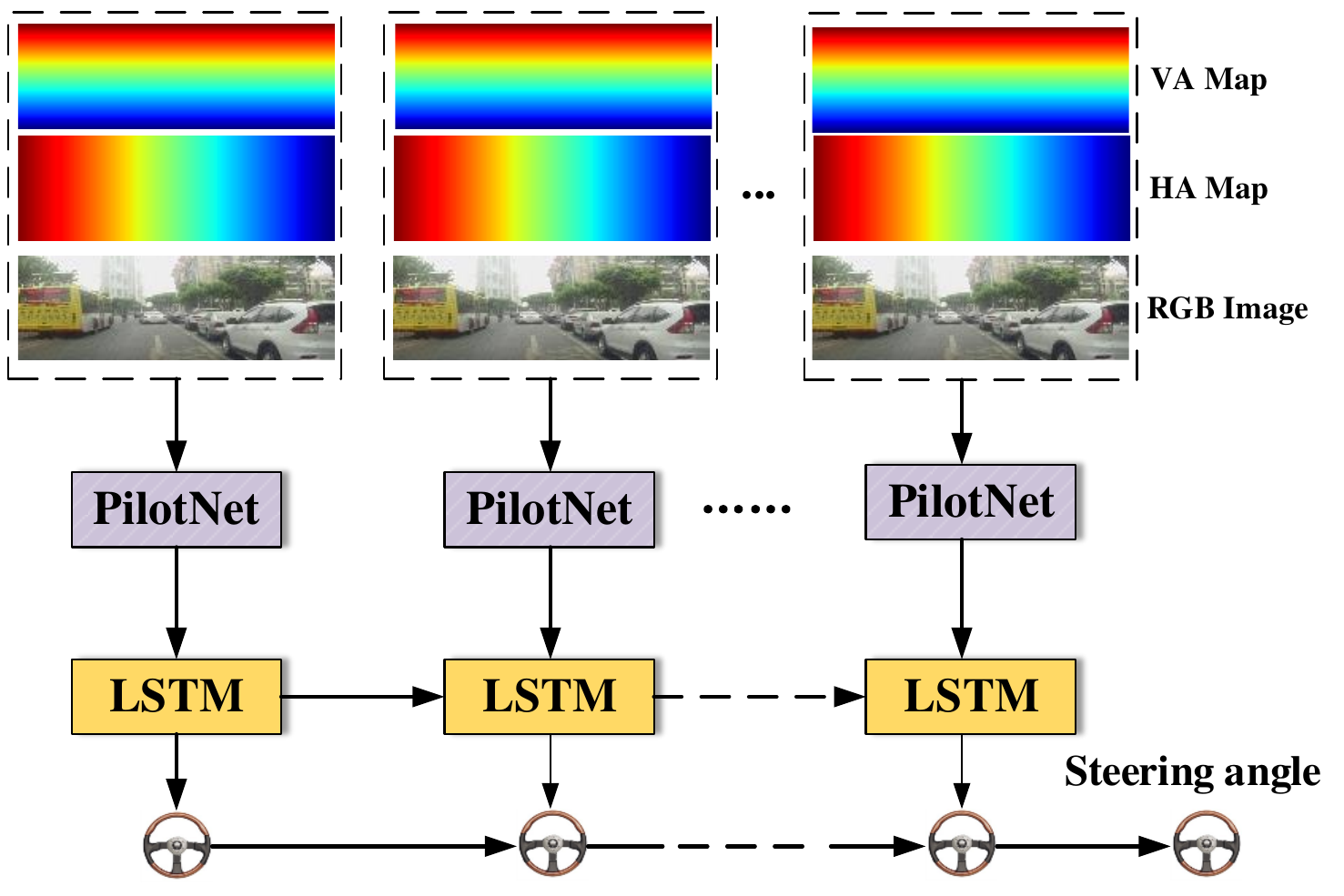}
	\caption{\textbf{Illustration of our framework.} Firstly, the PilotNet takes pixel-wise vertical angle maps (VA Map), horizontal angle map (HA Map) and RGB image as input to learn direction-aware features. Then, LSTM layers are exploit to obtain temporal cues and predict the current angle for steering control.}
	\label{fig:model}
\end{figure}

Following these network design strategies, we leverage the light-weighted PilotNet along with LSTM modules to estimate the steering angle. We did not adopt the previous steering wheel trajectory, due to the effects of the previous inaccurate steering angle estimate for current estimation. Consequently, our network can generate the temporal consecutive visual cues, which can be treated as the implicit environmental model. Predicting the steering angle can then further be formulated as a local path planning process,~\textit{i}.\textit{e}. lane keeping, and obstacle avoidance. Classical path planning algorithms make the decisions(going straight or turning)  based on a particular environmental representation, such as a divergent radial projection model for beam methods(BM)~\cite{fernandez2004bcm,lopez2019new} and an occupied grid model for artificial potential field methods~\cite{vadakkepat2000evolutionary}. These environmental models contain rich spatial information instead of simple binary semantic clues. On the contrary, visual features from deep neural networks have shown the capability of learning semantic cues in many tasks like image classification and detection~\cite{deng2009imagenet}. In order to perform privileged learning of spatial cues, we further introduce pixel-wise orientations of the inputting images to the network. Note that with these orientations, each pixel is a sector or radial beam, which makes the inputs alike the environmental model for BM. For a calibrated camera, pixel-wise orientations can be easily computed by the relation of perspective projection. Overall, the whole pipeline, learning end-to-end steering with direction-aware features, is shown in Fig. \ref{fig:model}.

Since deep features are learned automatically from a large amount of labeled data, the performance of the end-to-end steering driving depends on the quality of training data. In fact, the steering angles from diverse datasets often appear to imbalance distribution, namely, most of the steering values are close-zero and less of them have sharp values. Based on SteeringLoss~\cite{yuan2020steeringloss}, this paper proposes to solve the imbalance training problem by adding a gain factor to the ${\rm smooth}_{L_1}$ loss function for the backpropagation. In addition, we built a new end-to-end driving dataset to study the generalization ability of our model in different driving scenarios, such as urban, country, and off-road. Our dataset is collected using a camera-LiDAR combined product called Pandora~\footnote{\url{https://www.hesaitech.com/en/Pandora}}. Therefore, we named our dataset as PandoraDriving.

In summary, our main contributions are as follows:(i) we propose an end-to-end steering prediction model, that leverages temporal and spatial visual cues to produce direction-aware features; (ii) we make an improvement on the original SteeringLoss and partly solve the imbalance training problem; (iii) we present a dataset with more complicated driving scenarios than existing datasets.

\section{Related Works}
\subsection{Driving Model}

Nowadays, autonomous driving technology mainly realizes the perception of the environment and generates corresponding vehicle control commands. The autonomous driving framework can be divided into two main types: mediated perception approaches and end-to-end approaches. The mediated perception approaches \cite{chen2015deepdriving,gurghian2016deeplanes} detect and classify the surrounding environment to realize the mapping of the visual input to the pre-designed rules, and finally generates the corresponding vehicle control commands through the pre-designed rules. However, the mediated perception approaches rely on pre-designed rules, which are only applicable to some specific driving scenarios. In addition, the mediated perception approach is not suitable for complex situations, because it may miss some useful information for decision-making.

Different from the mediated perception approaches, the end-to-end approaches \cite{bojarski2018visualbackprop,bojarski2016end,kim2017interpretable,bojarski2017explaining,chi2017learning} directly map the visual input to the vehicle control command, without the need to manually design modules and rules. Pomerleau \cite{pomerleau1989alvinn} is the first one to use the end-to-end method in autonomous vehicle navigation. They built an autonomous driving system using a multi-layer perceptron (MLP)  and named Autonomous Land Vehicle in a Neural Network (ALVINN). But this network is very shallow, it can only drive in simple scenarios with few obstacles. The success of ALVINN demonstrates the potential of neural networks in autonomous driving. With the improvement of computing power and the development of deep learning, many researchers use convolutional neural network(CNN) to achieve environmental awareness and steering angle prediction. Nvidia \cite{muller2006off} first used the CNN to construct an autonomous driving system for robot in the DAVE project. This system can control the vehicles to avoid obstacles safely. After that, Nvidia built a PilotNet \cite{bojarski2016end} network on autonomous vehicles to predict the steering angle. This model performs well on simple highway roads with fewer obstacles. Yang \cite{yang2017feature} analyzed the influence of features on the performance of the end-to-end steering angle control model through experiments. The research demonstrates that road-related features are indispensable, and roadside-related features can improve the generalization of the model, but sky features do not contribute to steering angle prediction. However, one disadvantage of the above end-to-end work is that it only receives a single image as input, so it ignores the temporal relationship between image frames. Considering the temporal relationship between the image frames, Xu \cite{xu2017end} and Chen \cite{chen2018lidar} added LSTM to the network to improve the prediction accuracy of the steering angle. In this work, we use PilotNet as the feature extraction network. And in the same way, we add LSTM to the end-to-end steering angle control network, which is composed of CNN and LSTM. After receiving several continuous images as input, CNN network extracts the features of visual input, and feeds the extracted sequence features to LSTM for steering angle prediction.

Fernandez \cite{fernandez2004bcm} proposed a beam curvature method(BCM) method for obstacle avoidance of mobile robots in unknown environments. The BCM method combines beam method (BM) and curvature velocity method (CVM). In their work \cite{fernandez2004bcm}, the beam method first calculate the best one-step heading then pass it to CVM to calculate the optimal linear and angular velocities. In the local navigation of the mobile robot, the system obtains the divergent radial projection model of the environment through the beam method. Then the BM method calculates the best beam in the divergent radial projection model by applying the objective function, and then obtains the current best local heading. After that, Lopez \cite{lopez2019new} proposed a new local navigation approach for autonomous driving vehicles, which combined the Pure pursuit method and BCM. Therefore, in our autonomous vehicle system, end-to-end steering angle control model can also be considered as a local navigation system. In our work, we adopt the similar idea of BCM, which determines the best local navigation direction by building BM model. We first calculate the horizontal angle and vertical angle corresponding to each pixel, which are similar to the sector or radial beam in beam method, and then input them into the network. Then the network combines the visual input and the input of pixel-wise horizontal angle and vertical angle to predict the current local steering angle.

\subsection{Imbalanced Training}

For the end-to-end steering angle prediction model, imbalanced training is a great challenge at present. Since the car drives straight for most of the time, therefore, most of the steering angle is close to 0 in real driving data. As for imbalanced training, there are three ways to deal with this problem. One of the methods is to amplify the dataset, namely, increase the number of samples with the large steering angle. The main methods for dataset amplification are to recollect data or use data enhancement methods. Although amplifying the dataset is indeed an effective method, its cost is relatively high. In addition to amplify the dataset, Qing \cite{qing2019end} solves the problem of imbalanced training by re-sampling the dataset. In their work, they divided the steering angle into 199 bins, then randomly selected some samples from each bin and kept the number of samples in each box to no more than 2000. But this method may lose some sample data, especially when the dataset is small, some important scene data may be lost. The last method is to modify the loss function and apply different weights to different losses to enhance the influence of small sample data. As shown in the work of Wang \cite{wang2019end} and Yang \cite{yang2018end}, they apply different loss weight for different steering angle, the steering angle close to 0 is set a smaller weight while the larger steering angle is set a larger weight. But they did not give a specific loss function, it is difficult to determine the weights corresponding to different steering angle errors. In other words, it is difficult to distinguish between different steering angle errors. Aiming at this problem, Yuan \cite{yuan2020steeringloss} proposed a novel cost-sensitive loss function, which was inspired by Lin's idea \cite{lin2017focal}. They analyzed the data distribution of steering angle, and considered that the steering angle distribution was similar to the double long-tailed distribution. Finally, the SteeringLoss based on this distribution was proposed. The loss function can increase the influence of large steering angle while decrease the influence of small steering angle. With SteeringLoss, the model can estimate a wider range of steering angles. As SteeringLoss chooses mean square error (MSE) as the basic function, MSE is more sensitive to outliers and it is difficult to predict small steering angles. Therefore, we proposed our loss function and chose ${\rm smooth}_{L_1}$ as the basic function. With our loss function, the model can estimate the steering angle more accurately, and the predicted steering angle range is wider than SteeringLoss.

\subsection{Driving Datasets}
In order to train the end-to-end steering angle control model, some researchers have collected some real-life driving datasets, such as LiVi-Set\cite{chen2018lidar}, BDD 100K Dataset\cite{yu2020bdd100k}, Udacity Dataset. LiVi-Set contains video, point clouds and corresponding steering angle and speed values. However, camera calibration data is not included in the data, which is used to calculate the horizontal angle and vertical angle. The main driving scenarios included in the dataset are urban roads and highways. The BDD dataset contains only video, GPS and IMU, and corresponding steering angle values is not collected. The steering angle values is calculated by IMU information, so the accuracy of steering angle can not be verified. Udacity Dataset mainly contains the image of the highway and the corresponding steering angle data, but does not contain the image calibration data. Our dataset contains images, camera calibration data, point clouds and corresponding throttle position, steering angle and brake position, and includes multiple driving scenarios, such as urban, country, and off-road. The off-road driving scene is not available in many datasets.

\section{Methods}
\subsection{End-to-end Driving Model Combined with Pixel-wise Orientations}

\begin{figure}
	\centering
	\includegraphics[width=1\linewidth]{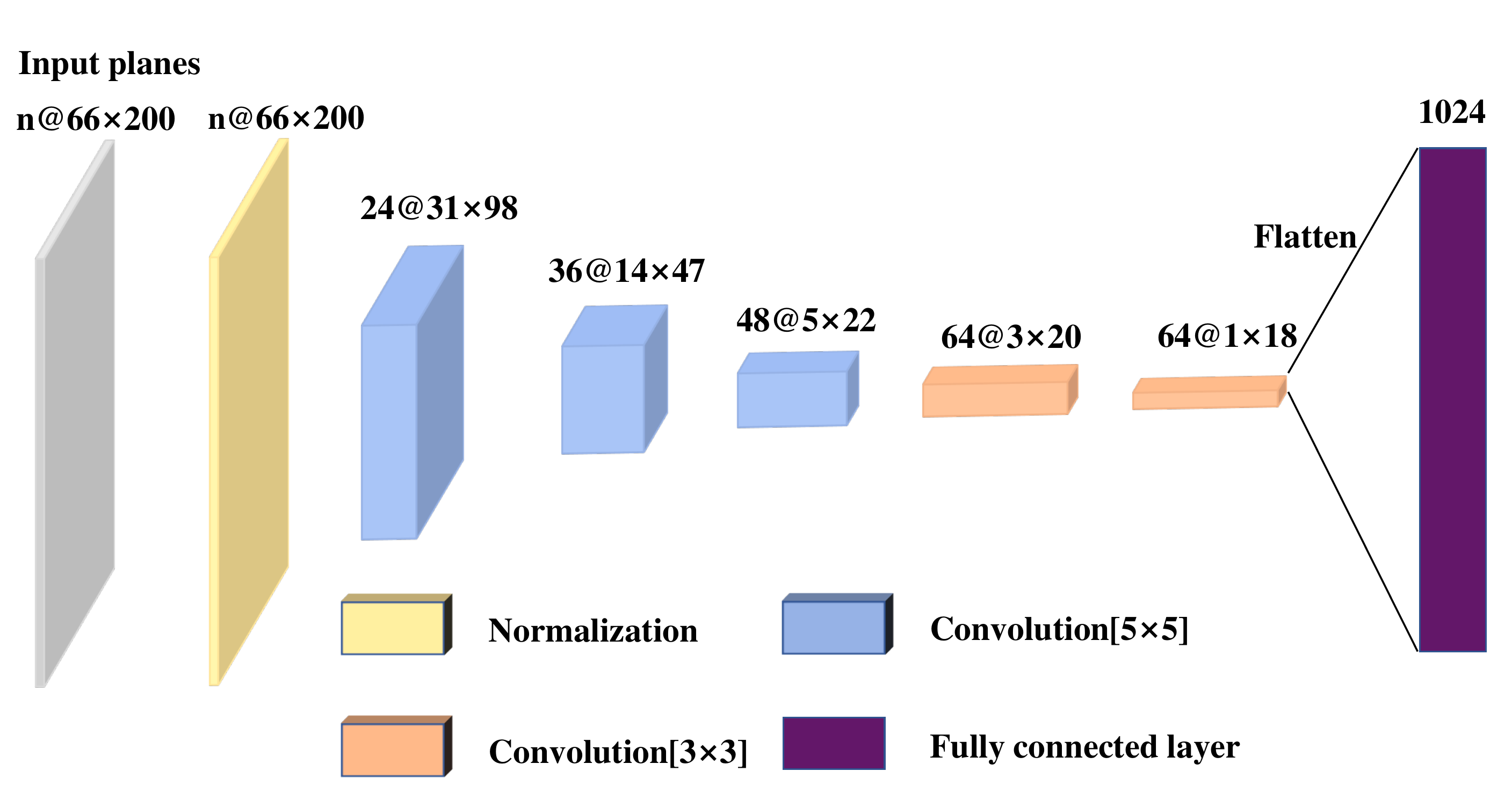}
	\caption{\textbf{The PilotNet architecture.} The PilotNet is used as a feature-extracting sub-network to generate  image features and forward the extracted features to LSTM. $n$ represents the channel of the input image.}
	\label{fig:pilotnet}
\end{figure}

From previous work, it can be concluded that the end-to-end steering angle prediction model is difficult to generate the accurate steering angle only based on one single image. In the real driving scene, the driver's driving behavior is continuous. Therefore, similar to human driving behavior, many end-to-end steering angle control models use continuous frames as input. On the one hand, the CNN model can only extract the spatial features of a single image, so it will ignore the temporal relationship between frames; on the other hand, as a special form of RNN, LSTM can extract the temporal information of sequence data. Therefore, we propose our end-to-end steering angle control model, which combines the CNN network and the LSTM time series network. The overall network architecture is shown in Fig. \ref{fig:model}. In addition, considering that the size of our training data is small, we use the light-weighted Pilotnet as the spatial feature extraction layer for the image sequence. The PilotNet architecture is shown in Fig. \ref{fig:pilotnet}. In this work, we use the first 6 layers of PilotNet to encode visual features, which includes 5 convolutional layers and 1 fully connected layer. The first normalization layer normalizes the image and speeds up the convergence of the network. The convolutional kernel size of the first three convolutional layers is $5\times5$ and the stride size is $2\times2$, while the last two convolutional layers adopt a $3\times3$ kernel and a $1\times1$ stride. Then a fully connected layers of sizes 1024 follows the convolutional layers. After PilotNet extracts the visual features, the encoded features are forward to the LSTM layer. As seen in the whole pipline of the entire network, the driving model takes the image sequence as input and encodes the visual input through the PilotNet, where the image sequence includes the RGB image and the pixel-wise horizontal angle map(HA Map) and vertical angles map(VA Map). Then the encoded visual features are fed into the LSTM layer to encode the temporal information. Finally, the network predicts the steering angle according to the features encoded from LSTM.

\subsection{Pixel-wise Orientations}

As we mentioned above, our end-to-end steering angle control model can be simply regarded as local navigation. Most end-to-end control models use images as visual input, which helps the end-to-end network to understand the surrounding scene and then realize the mapping from pixels to steering angles. But in our work, we first calculate the pixel-wise orientations, which include the corresponding horizontal angle map and vertical angle map for each pixel in the image, and then concatenate the horizontal angle map and vertical angle map with the image into a five-channel image. Finally, the synthesized image is fed into the end-to-end driving model as input. By adding the pixel-wise orientations, the end-to-end steering angle control network can better understand the heading angle of objects (such as cars, pedestrians, lane lines, etc.) relative to the autonomous vehicle. With the relevant heading angle information, the end-to-end network can avoid predicting the angle with obstacles as much as possible when predicting the steering angle.

\begin{figure}
	\centering
	\includegraphics[width=0.7\linewidth]{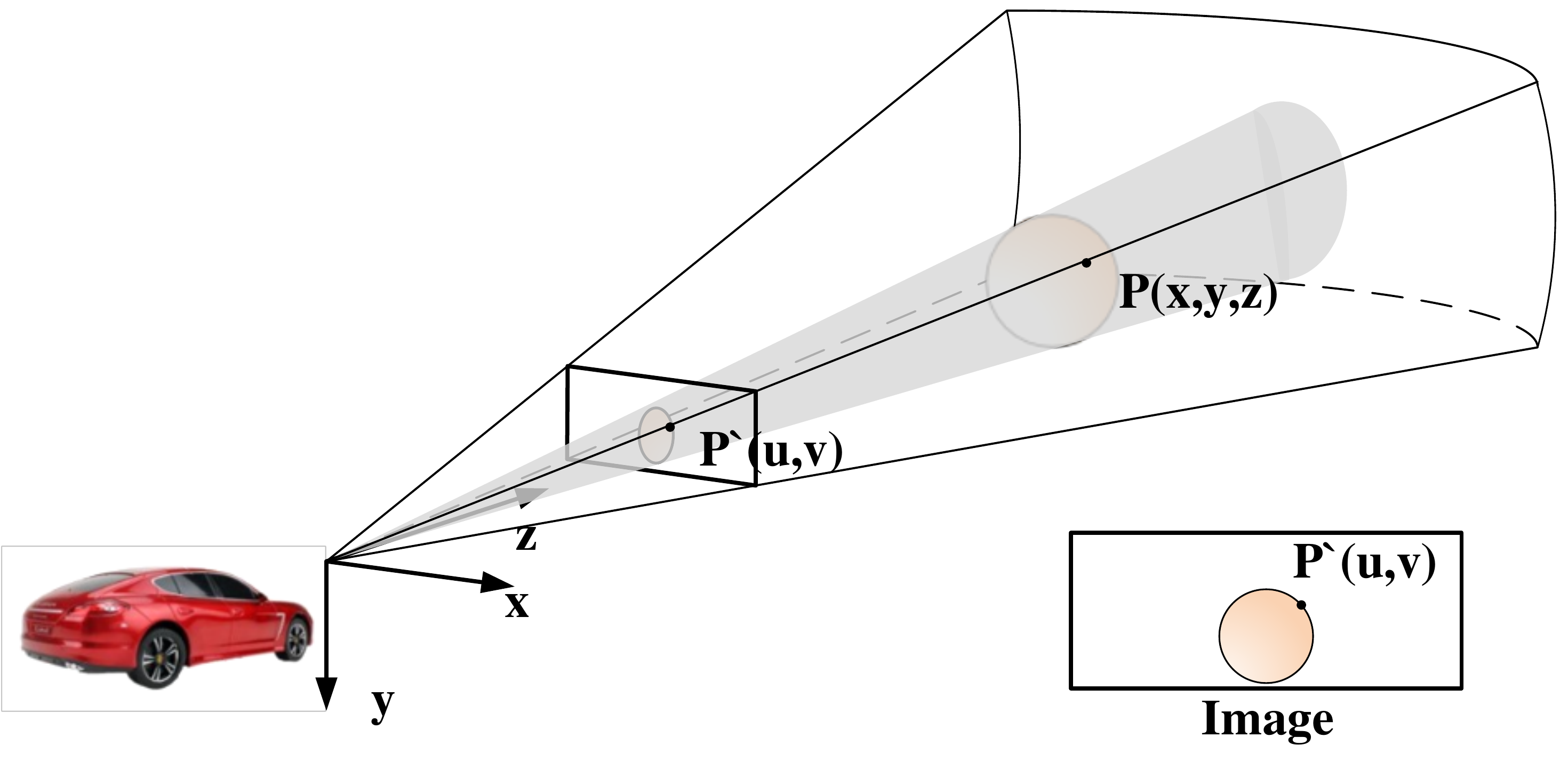}
	\caption{\textbf{Local navigation scene.} Our end-to-end driving model can be treated as local navigation. The local navigation scene is simplified as a frustum under the field of view angle of the image, which is in the camera coordinate system. For convenience of expression, we use a ball to represent obstacles in the scene. There is a point $P$ on the ball, which is in the camera coordinate system, and its imaging point $P'$ in the image coordinate system.}
	\label{fig:2}
\end{figure}

\begin{figure}
	\centering
	\includegraphics[width=0.8\linewidth]{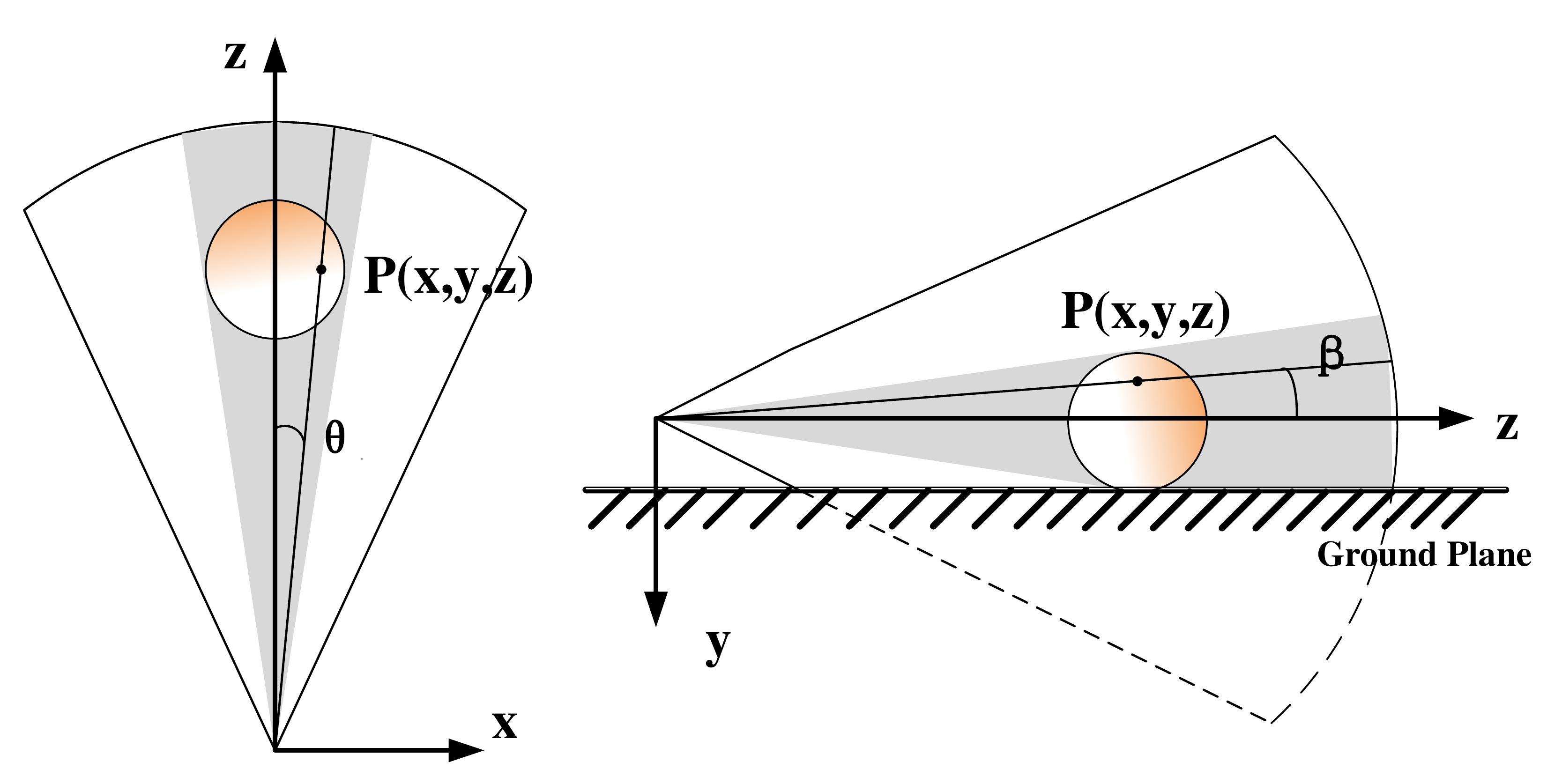}
	\caption{\textbf{Projection of local navigation scene.} The image on the left is projected along the positive direction of the y-axis, and the image on the right is projected along the negative direction of the x-axis.}
	\label{fig:hor_ver}
\end{figure}

Our driving scene can be simplified as shown in Fig. \ref{fig:2}. As shown in Fig. \ref{fig:2}, the scene under the field of view of the image is described by a frustum, which is in the camera coordinate system. In order to describe the driving scene conveniently in the image, we simplify the scene, which contains only one ball used to represent objects. In this simplified scene, we assume that there is a point $P $ on the ball, and its imaging point in the image is $P'$. Afterwards, we project the simplified scene along the positive direction of the y-axis and the negative direction of the x-axis to obtain two projections in Fig. \ref{fig:hor_ver}. As shown in Fig. \ref{fig:hor_ver}, $\theta$ and $\beta$ are the horizontal angle and vertical angle of point $P$ respectively. In Fig. \ref{fig:hor_ver}, the collision area indicated by the gray area. If the autonomous vehicle travels in the gray area, it may collide with the object. Therefore, for the local navigation of autonomous vehicles, the recognition of gray areas can help autonomous vehicles avoid obstacles. If the horizontal angles and vertical angles corresponding to the entire image are fed into the end-to-end steering angle control model, the model can identify the corresponding gray area through learning, and then predict the steering angle more accurately.

After simplifying the scene, we need to calculate the horizontal angle and vertical angle of point $P $ in Fig. \ref{fig:hor_ver}, which are also the pixel-wise orientation of point $P'$. According to the camera imaging principle, the relationship between the point $P$ and the pixel point $P'$ can be expressed by the following formula:
\begin{equation}\label{form1}
u=f_x\times\frac{x}{z}+c_x 
\end{equation}
\begin{equation}\label{form2}
v=f_y\times\frac{y}{z}+c_y
\end{equation}

where $c_x$, $c_y$, $f_x$, and $f_y$ are camera internal parameters. After that, we can get the following formulation from Fig. \ref{fig:hor_ver}:
\begin{equation}\label{form3}
tan\theta=\frac{x}{z}
\end{equation}
\begin{equation}\label{form4}
tan\beta=\frac{y}{z}
\end{equation}
Combining the above four equations, we can calculate the horizontal angle and vertical angle corresponding to each pixel by the following formulation:
\begin{equation}\label{form5}
\theta=atan2(u-c_x,f_x)
\end{equation}
\begin{equation}\label{form6}
\beta=atan2(v-c_y,f_y)
\end{equation}

\subsection{Loss Function}
\begin{figure}
	\centering
	\subfigure{
		\includegraphics[width=0.47\linewidth]{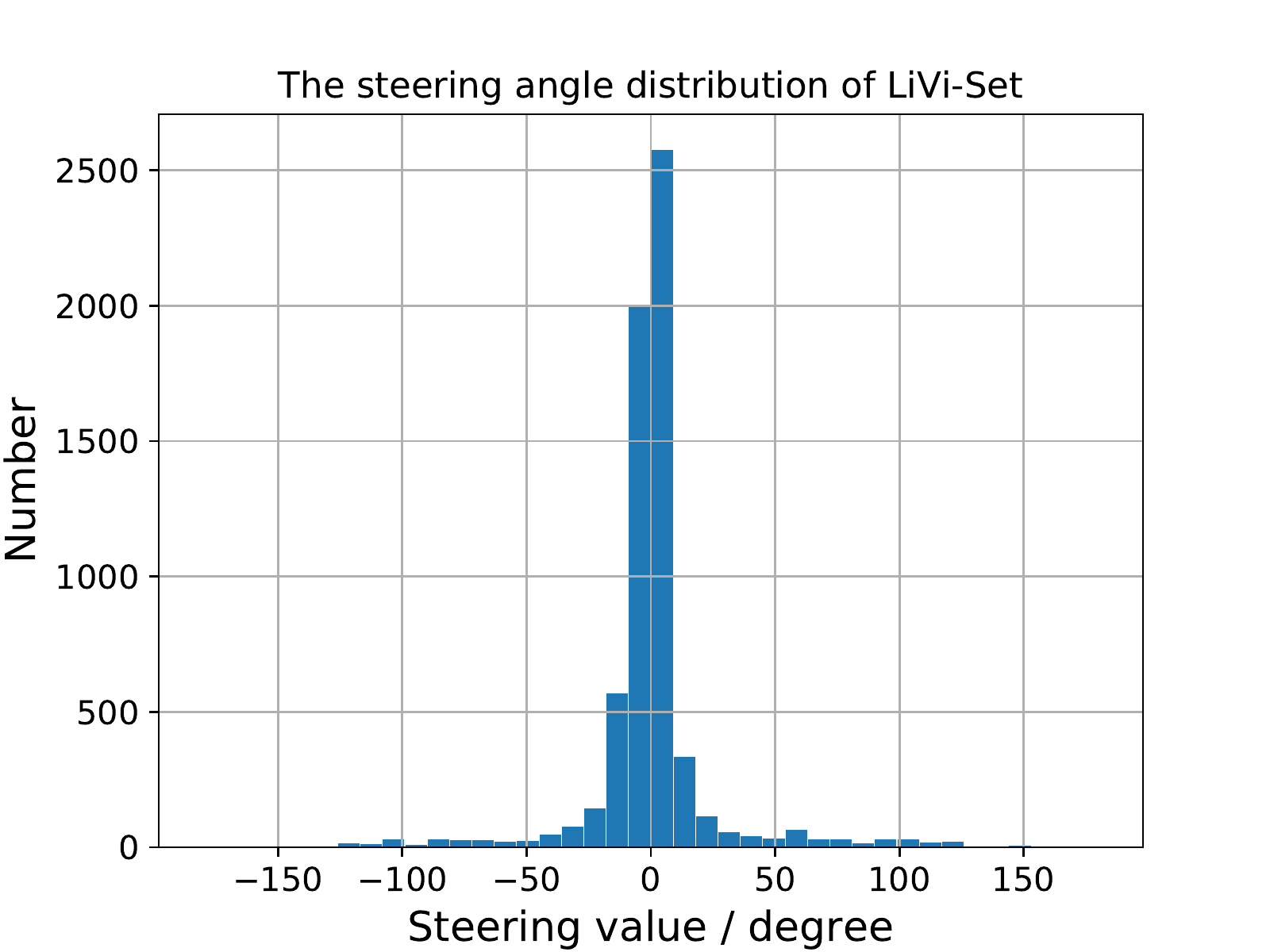} 
	}
	\subfigure{
		\includegraphics[width=0.47\linewidth]{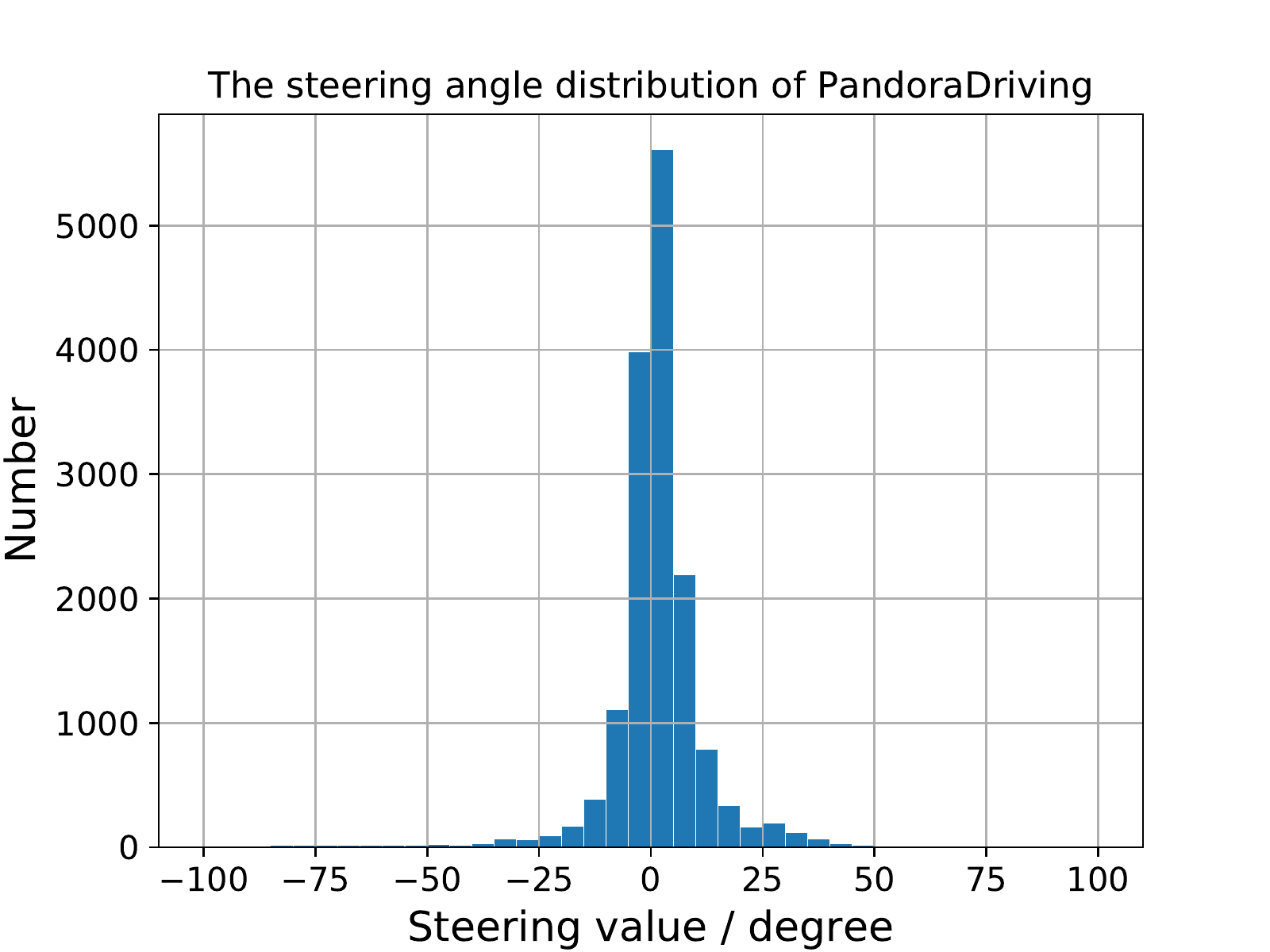}
	}
	\caption{\textbf{The steering value Distribution.} This figure demonstrates the steering value distribution of LiVi-Set and PandoraDriving. The steering angle of left turn relative to standard angle is less than 0, and that of right turn relative to standard angle is greater than 0.}
	\label{fig:distribution}
\end{figure}

Since the existing relevant datasets are imbalanced, it is a challenge to train the end-to-end steering angle control model. Before introducing the loss function, we first analyze the distribution of the imbalanced dataset, include LiVi-Set and our dataset. The distribution of these two datasets is shown in Fig. \ref{fig:distribution}. From Fig. \ref{fig:distribution}, we can see that there are some similarities in the distribution of steering angle between the two datasets, their data distribution is similar to the long tail distribution. In both datasets, most steering angle values are concentrated near zero, and the larger the steering angle, the smaller the number of samples. If the imbalanced dataset is used to train the end-to-end steering angle control network, the network will tend to predict the smaller steering angle due to the influence of large sample steering angles, which are close to 0.

In order to solve the imbalanced training problem, we adopt the same basic principles as \cite{yuan2020steeringloss,wang2019end,yang2018end}. The basic principle is that small steering angles are assigned small weights, and large steering angles are assigned large weights. In this way, the model can focus more on large steering angles. Therefore, Yuan's \cite{yuan2020steeringloss} proposed a cost-sensitive loss function, which called SteeringLoss. By using SteeringLoss as the loss function, the end-to-end model can estimate a wider range of steering angles. In our work, we use modified the SteeringLoss and propose our loss function, which is called SteeringLoss2. The SteeringLoss is defined as follows:
\begin{equation}\label{form7}
loss=\frac{1}{2n}\sum_{i=1}^n(1+\alpha|y_i|^\gamma)^\delta (y_i-y'_i)^2
\end{equation}
where $n$ is the total number of samples, $y_i$ is the ground truth of $i^{th}$ sample, and $y'_i$ is estimation value of $i^{th}$ sample, $\alpha$, $\gamma$, $\delta$ are hyperparameters. SteeringLoss is based on mean square error (MSE), which is sensitive to outlier. Besides, when there is a big difference between the estimated value and ground truth, MSE is easy to cause gradient explosion. In the end-to-end steering angle control model, the MSE based model is difficult to estimate the small steering angle accurately because the MSE is more sensitive to the large steering angle. In the MSE based SteeringLoss, although the SteeringLoss can pay more attention to the large steering angle by assigning a large weight to the large steering angle, it is difficult to predict the small steering angle due to the influence of the basic function MSE. Therefore, we modify SteeringLoss and design our loss function as follows:
\begin{equation}\label{form8}
loss=\frac{1}{n}\sum_{i=1}^n(1+\alpha|y_i|^\gamma){\rm smooth}_{L_1}(|y_i-y'_i|)
\end{equation}
in which
\begin{equation}\label{form9}
{\rm smooth}_{L_1}(x)=
\begin{cases}
0.5x^2, \quad |x|\leq 1 \\
|x|-0.5,\quad otherwise
\end{cases}
\end{equation}
where $n$ is the total number of samples, $y_i$ is the ground truth of $i^{th}$ sample, and $y'_i$ is estimation value of $i^{th}$ sample. In the SteeringLoss2, we discard the hyperparameter $\delta$, because by setting the hyperparameter $\gamma$, the same effect as the hyperparameter $\delta$ can be achieved. The Steeringloss2 is a cost-sensitive loss function based on ${\rm smooth}_{L_1}$, while SteeringLoss is based on MSE. The reason why we choose ${\rm smooth}_{L_1}$ as the basic function is that ${\rm smooth}_{L_1}$ not only has the robustness of mean absolute error (MAE) but also can solve the gradient explosion problem of MSE. In the back propagation of neural network, since the update gradient of MAE remains unchanged, the model is difficult to converge when the loss is small. When the difference between the estimation value and the ground truth is small, ${\rm smooth}_{L_1}$ uses MSE, which can solve the problem of constant update gradient of MAE. When the difference between the estimation value and the ground truth is large, the MAE is used in ${\rm smooth}_{L_1}$. Although the end-to-end steering angle control model based on ${\rm smooth}_{L_1}$ predicts the small steering angle, the model based on SteeringLoss2 pays more attention to the large steering angle by assigning large weights to large steering angles. Therefore, the model based on SteeringLoss2 can predict both small steering angle and large steering angle.

\section{Experimental}
\subsection{Dataset Description}
In our work, LiVi-Set Prepared Dataset and our own collected datasets are used to verify our proposed method. The LiVi-Set contains point clouds, video, and corresponding driving behavior data, where the video is captured by the minor distortion dashboard camera. This raw dataset they provided has not been preprocessed and there are some video and driving behavior data misalignments, which will lead to poor performance of the model. In order to reduce the impact of noise on the performance of the model, we use their prepared dataset with few errors, although the prepared dataset size is small. In addition, the image resolution in the LiVi-Set is $1920\times1080$, while the input image resolution of our model is $200\times66$, so we need to preprocess the original data. We first cropped the image area related to the sky and the hood of the autonomous car, and then scaled it to the size of $200\times66$.

The driving scenarios in the LiVi-Set prepared dataset are mostly urban roads and highways. In order to verify whether our method is suitable for a variety of driving scenarios, we collected our own dataset through our own experimental platform and named it Pandoradriving. The experimental platform is equipped with Pandora lidar camera sensing kit made in China, which can simultaneously collect point clouds of 40 lines laser around the platform, color high-definition images in front of the vehicle, panoramic black-and-white images. Basides visual information, we also collect the driving behaviors through the Controller Area Network (CAN) bus. The collected driving behaviors include throttle position, steering angle, and brake position. In order to align the driving behavior data with the visual input, we adopt the network time alignment on the experimental platform, in which Pandora and CAN bus work at 10HZ.

\begin{figure}[ht]
	\centering
	\includegraphics[width=1\textwidth]{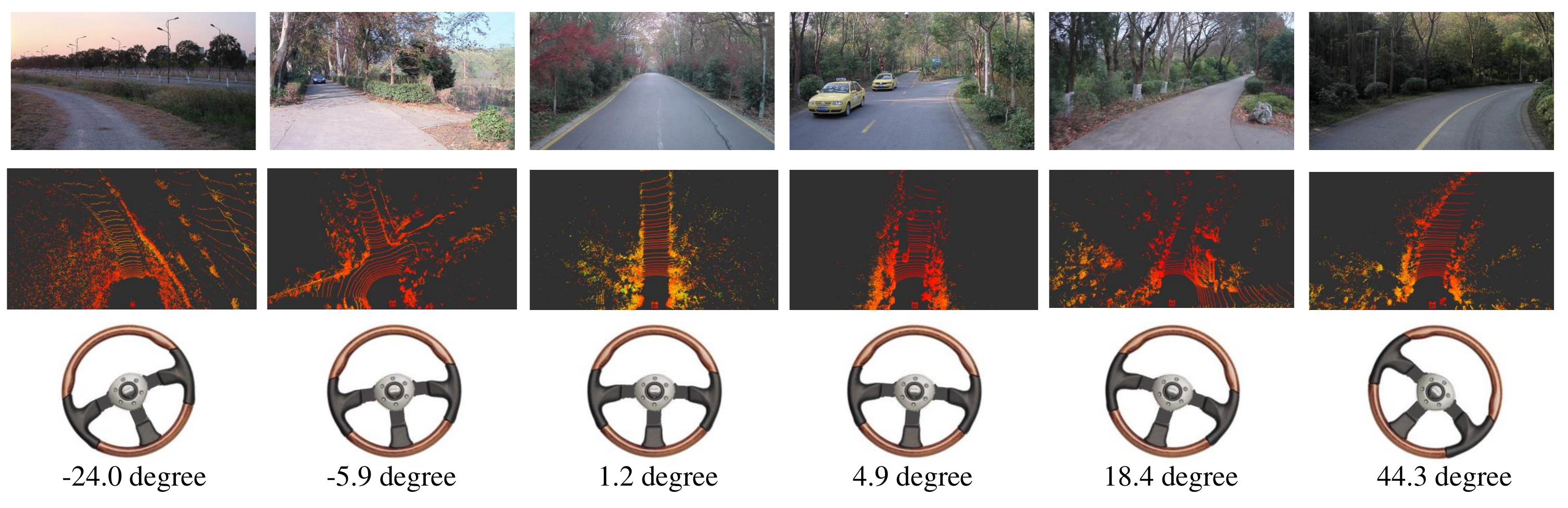}
	
	\caption{\textbf{The representative examples in PandoraDriving.} Examples include RGB images, point clouds, and corresponding steering angles in multiple driving scenarios, where driving scenarios include cities, rural areas, and suburbs. First row is the RGB image, second row is the corresponding point clouds and the last row is the corresponding steering angle.}
	\label{fig:exmple}
\end{figure}

Our dataset contains 18 different color image sequences, point clouds sequences, and corresponding behavior data. The total collection time is more than 40 minutes, and the total number of images collected is more than 25000 frames. Our dataset includes multiple driving scenarios,such as urban, country and off-road. As shown in Fig. \ref{fig:exmple}, we extracted some representative data from PandoraDriving, including corresponding images, point clouds, and steering angles in different scenes.

The main goal of our work is to train the end-to-end steering angle control model, so we only use images and corresponding driving behavior data. Since the network for image is $200\times66$, and the image resolution in our dataset is $1280\times720$, the image needs to be cropped and scaled before feeding into the network. We use the same processing method as LiVi-Set, first cropped the image area of the car hood and the sky, and then scaled it to the specified size. When the steering wheel turns left relative to the standard angle, the steering wheel value is less than 0. Similarly, the steering wheel greater than 0 indicates that the steering wheel turns right relative to the standard angle. In addition, the range of the steering angle collected in the dataset is $[-\pi\times1000,\pi\times1000]$, and the corresponding real steering angle is $[-\pi,\pi]$, so we need to scale all the raw steering angles to $[-\pi,\pi]$.

\subsection{Evaluation Metric}
In order to evaluate the performance of the end-to-end steering angle prediction model, the researchers proposed different evaluation metric. In PilotNet, they used the calculation of the number of manual interventions as evaluation metric in the simulation test, while in the on-road test, the proportion of autonomous driving time was used as evaluation metric. However, the evaluation metric cannot distinguish the different levels of bad predictions. In other words, we cannot distinguish the difference between the estimated steering angle and the ground truth by 5 degrees and 15 degrees. Xu \cite{xu2017end} proposed a driving perplexity as an evaluation criterion, which is inspired by language modeling metric. However, this evaluation metric does not give meaning to the real world, nor can it judge whether the model is effective. In other words, it is difficult to judge the specific difference between the predicted value and the real value from the evaluation metric. Accuracy metric is widely adopted in \cite{chen2018lidar,muller2006off,kim2017end}, and the accuracy metric is more intuitive than the driving perplexity. Therefore, we also adopt the accuracy metric in our work. In order to facilitate the calculation accuracy, we need to use the tolerance threshold. We believe that in the real world, the driver's driving behavior also has a small deviation, so it is reasonable to use the tolerance threshold. In our work, We chose {$5^\circ$} as the steering angle tolerance threshold, while Chen \cite{chen2018lidar} choose $6^\circ$ as the the steering angle tolerance threshold. In addition to using accuracy as the evaluation criterion, we also use the standard deviation (SD) of the estimated value to evaluate the model, which is also used in \cite{yuan2020steeringloss}. SD is used to evaluate the value range of the steering angle estimated by the model. The larger the SD of the estimated steering angle, the wider the steering angle can be predicted by the model. In other words, if the SD of the steering angle estimated by the model is larger, it indicates that the model can estimate the larger steering angle instead of only estimating the steering angle close to 0.

\subsection{Results}
\begin{figure}[ht]
	\centering
	\subfigure[LiVi-Set]{
		\includegraphics[width=0.9\textwidth]{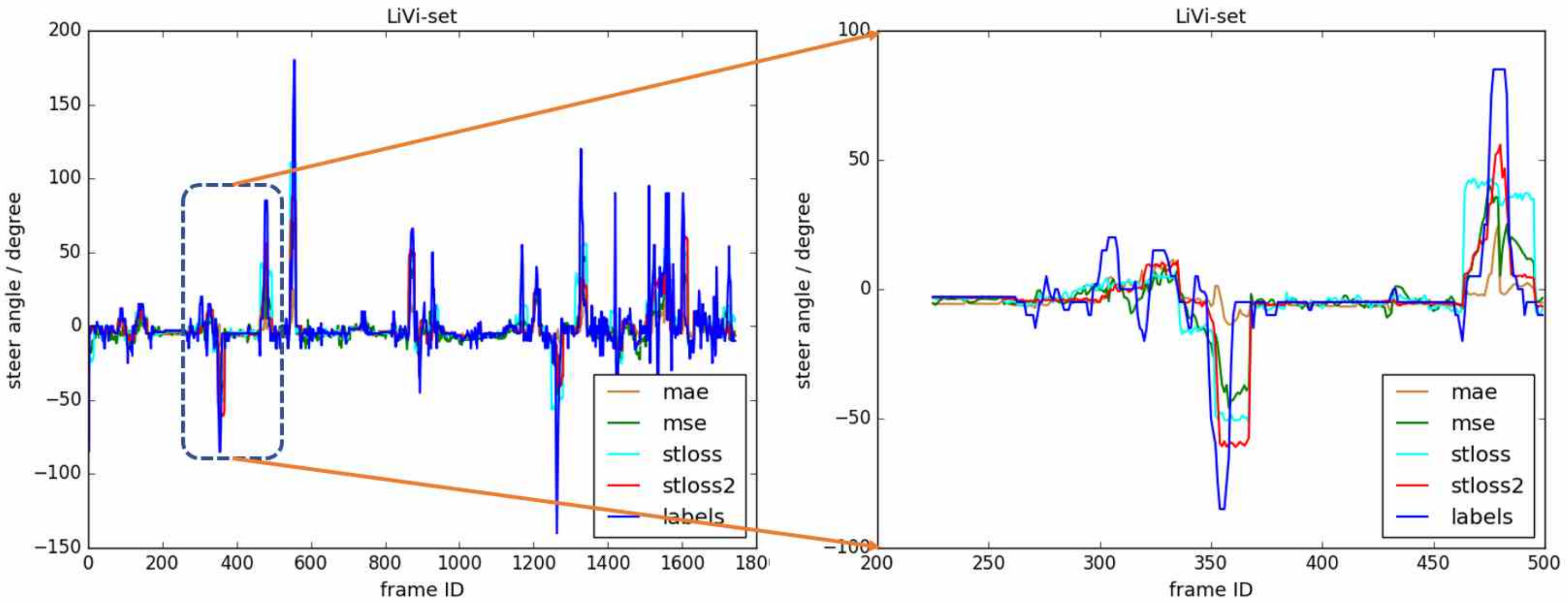}
	}
	\subfigure[PandoraDriving Dataset]{
		\includegraphics[width=0.9\textwidth]{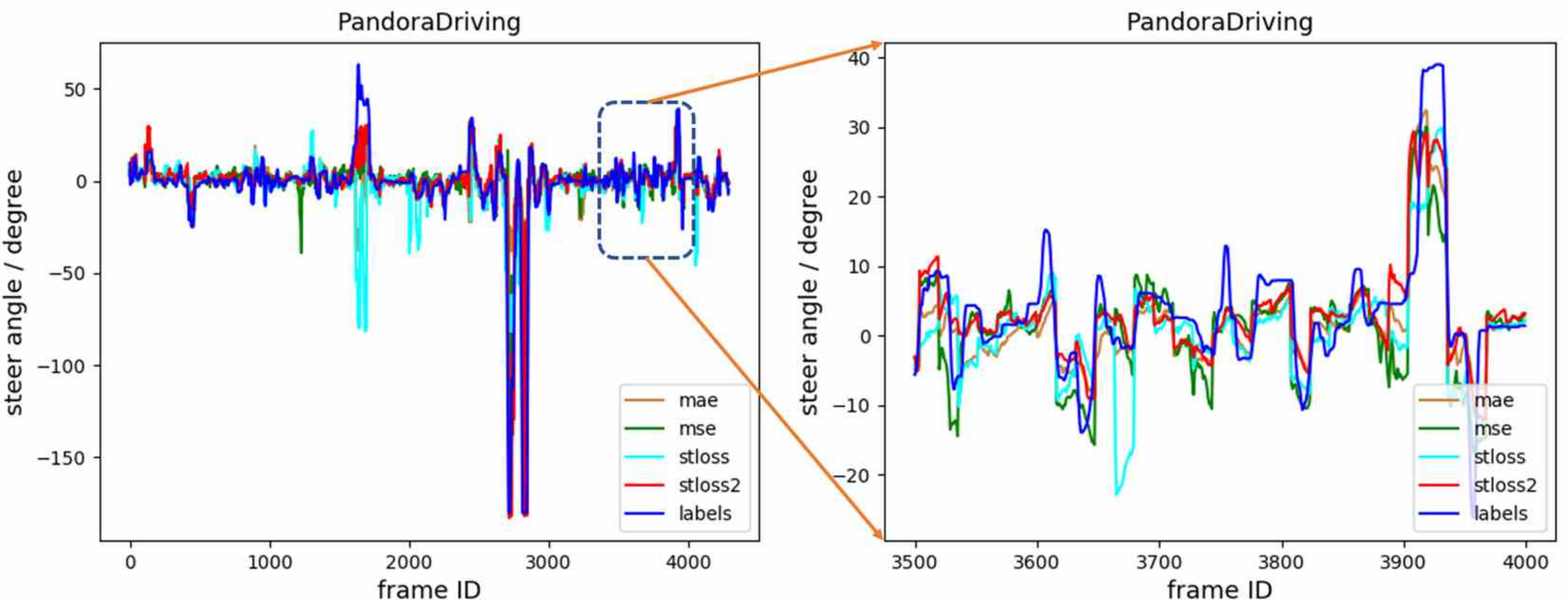}
	}
	\caption{Comparison of the ground truth and the predicted steering of the network under different loss functions. In the figure, stloss is the model with SteeringLoss, stloss2 is the model with SteeringLoss2, and label is the ground truth. The right sub-figure is the sub-sequence of the left sub-figure such that more detailed difference can be clearly observed.}
	\label{fig:plot0}
\end{figure}

In order to verify the impact of the SteeringLoss2 on the performance of the model, we did four comparative experiments using MAE, MSE, SteeringLoss and SteeringLoss2 as the loss function. All experiments are based on the model shown in Fig. \ref{fig:model}. At this time, the input of the model is the RGB image sequence, and the structure of the network feature extraction module is shown in Fig. \ref{fig:pilotnet}. The steering angle prediction accuracy of our experiments is shown in Table \ref{table:1}, where the steering angle tolerance threshold is set to 5 degrees, and the parameter $\alpha$, $\gamma$ and $\delta$ are set to 1.0. To further investigate the performance of the model under different loss functions, Fig. \ref{fig:plot0} plots the ground truth of consecutive frames collected from human and the corresponding predicted steering angle values. In addition, in order to observe the difference between the predicted value and the ground truth in more detail, some consecutive frames are selected from the left sub-figure for display, and the results are shown on the right sub-figure in Fig. \ref{fig:plot0}. As shown in Table \ref{table:1}, the model using SteeringLoss2 as the loss function has the highest prediction accuracy, while the model using MSE as the loss function has the lowest prediction accuracy. MSE calculates the square of the distance between the ground truth and the estimation, while MAE only calculates the absolute error between the ground truth and the estimation. In other words, MSE is more sensitive to outliers than MAE. Therefore, in imbalanced training, due to the higher proportion of steering angles close to 0 degrees, the steering angles predicted by MAE are relatively close to 0, and it is difficult to predict large steering angles. As can be seen from the prediction results of LiVi-set in Fig. \ref{fig:plot0}, the steering angle predicted by MAE between 340 and 380 frames is closer to zero than other loss functions. On the contrary, the model using MSE as the loss function is vulnerable to the influence of large steering angle, which makes the predicted steering angle very unstable, and the prediction accuracy of small steering angle with high proportion is worse than MAE. Therefore, the prediction accuracy of the model with MAE loss function is higher than that of MSE loss function which is sensitive to large steering angle. It can be seen from Table \ref{table:1} that the prediction accuracy of the model using the SteeringLoss or SteeringLoss2 is higher, which also proves that the cost-sensitive loss function is effective in the imbalanced training. Compared with the traditional loss function, this cost-sensitive loss function can improve the model's attention to large steering angle by applying larger weight to large steering angle and smaller weight to small steering angle, so as to maintain the prediction accuracy of small steering angle and improve the prediction of large steering angle. Observing the impact of the two cost-sensitive loss functions on the performance of the model, it can be seen that the prediction accuracy of the improved loss function SteeringLoss2 is better than that of the SteeringLoss. Furthermore, the proposed SteeringLoss2 choose ${\rm smooth}_{L_1}$ as the basic function, which has the robustness of MAE, and at the same time will not be too sensitive to outliers like MSE. Therefore, the prediction accuracy of SteeingLoss is higher than MSE, while SteeringLoss2 performs best. In addition to the observation accuracy, it can be observed from SD that the SD of the steering angle predicted by the model using MAE as the loss function is the smallest. The reason is that MAE is affected by the small steering angle, which accounts for a high proportion of the number of samples, and its predicted steering angle is small, so it cannot cope with the scene of large steering angle, which leads to the smallest SD value. On the contrary, the model with cost-sensitive loss function can get larger SD than the traditional loss function. In addition, it can be found that the model with SteeringLoss as the loss function on the LiVi-set has a larger SD value, but combining with Fig. \ref{fig:plot0}, it can be seen that most of the steering angles predicted by SteeringLoss are larger than the ground truth, so the SD value is large and the accuracy is low. When SteeringLoss2 is used as the loss function on PandoraDriving, the model gets the largest SD, which also confirms that the cost-sensitive loss function SteeringLoss2 can predict a relatively large steering angle, and the predicted steering angle is also more accurate.

\begin{table}[ht]
	\centering
	\caption{\textbf{Performance of models with different loss functions. } Accuracy represents the prediction accuracy of the steering angle of the model. SD represents the  standard deviation of the  steering estimation value. The accuracy is measured within $5^\circ$ biases.}
	\label{table:1}
	\setlength{\tabcolsep}{5mm}{
		\begin{tabular}{ccccc}
			\hline 
			\multirow{2}{*}{loss function} & \multicolumn{2}{c}{LiVi-Set } & \multicolumn{2}{c}{PandoraDriving} \\
			\hhline{~----}
			&Accuracy & SD  & Accuracy & SD  \\
			\hline 
			MAE	& 66.3\% &5.34  & 72.2\% &9.58 \\ 
			
			MSE	& 56.3\% & 12.97 &  69.4\% &15.34 \\ 
			
			SteeringLoss & 61.8\%&16.37 & 69.7\% &15.57\\ 
			
			SteeringLoss2 & 67.0\%&14.73  & 76.4\% &22.99 \\ 
			
			\hline 
	\end{tabular}}
\end{table}

\begin{figure}[]
	\centering
	\subfigure[LiVi-Set]{
		\includegraphics[width=0.9\textwidth]{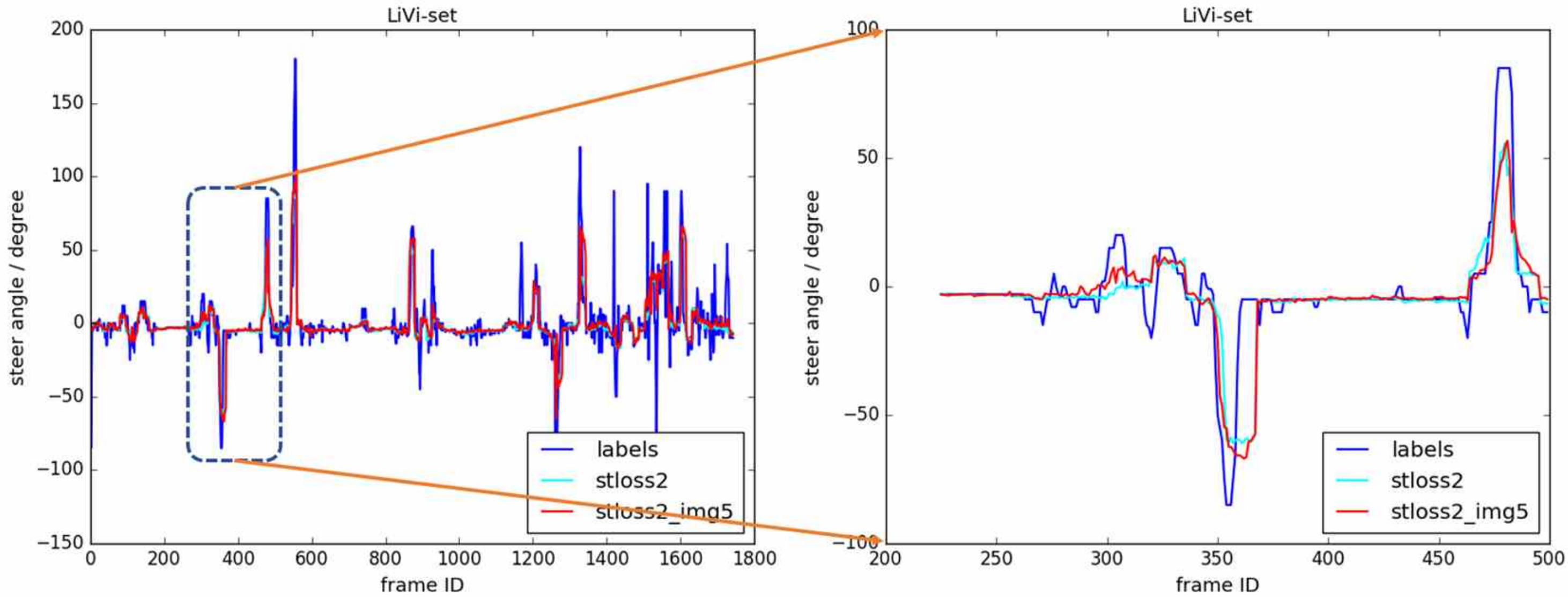}
	}
	\subfigure[PandoraDriving Dataset]{
		\includegraphics[width=0.9\textwidth]{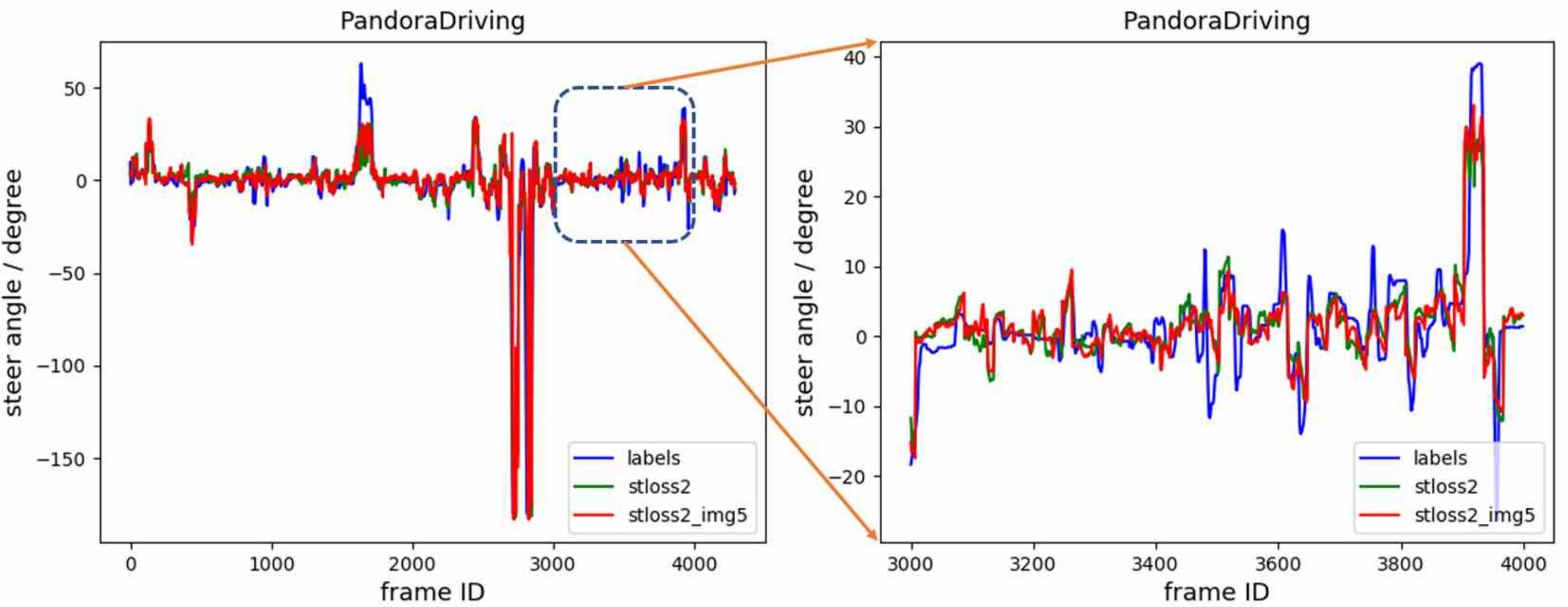}
	}
	\caption{Comparison of the ground truth and the predicted steering of the network under different inputs. In the figure, stloss is the model with SteeringLoss, stloss2 is the model with SteeringLoss2, and label is the ground truth. The right sub-figure is the sub-sequence of the left sub-figure such that more detailed difference can be clearly observed.}
	\label{fig:plot1}
\end{figure}

In order to verify the influence of pixel-wise orientations on network performance, we concatenate the pixel-wise orientations with RGB image matrix into five-channel input matrix. In addition, it can be seen from Table \ref{table:1} that the accuracy and SD of the cost-sensitive loss function used by the network are better than the traditional loss function. Therefore, the comparison experiment here uses cost-sensitive loss functions, and compares the effect of pixel-wise orientations on the performance of the model through experiments. The model performance are shown in Table \ref{table:2}, and the comparison performance between the visualized predicted steering angle and the ground truth are shown in Fig. \ref{fig:plot1}. As shown in Table \ref{table:2}, after adding the pixel-wise orientations as an additional input, the accuracy of the network and the SD of the predicted steering angle are higher than those of the network using only RGB images as input. In addition, it can be seen from Fig. \ref{fig:plot1} that the steering angle predicted by the network is closer to the ground truth after adding the pixel-wise orientations. This proves that the network can recognize the object in the image and predict more accurate steering angle information through the learning of direction-aware features. At the same time, the standard deviation of the predicted steering angle is slightly larger, which indicates that the model can predict large steering angle.

\begin{table}[ht]
	\centering
	\caption{\textbf{Performance of models with different input. } Accuracy represents the prediction accuracy of the steering angle of the model. SD represents the  standard deviation of the  steering estimation value. The accuracy is measured within $5^\circ$ biases.}
	\label{table:2}
	
	\begin{tabular}{ccccc}
		\hline 
		\multirow{2}{*}{Input + loss function} & \multicolumn{2}{c}{LiVi-Set } & \multicolumn{2}{c}{PandoraDriving} \\
		\hhline{~----}
		&Accuracy & SD  & Accuracy & SD  \\
		\hline 
		RGB + SteeringLoss2 & 67.0\% & 14.73  & 76.4\% & 22.99 \\ 
		
		RGB + pixel-wise orientations + SteeringLoss2 & 67.9\% & 15.12  & 77.9\% & 25.52 \\ 
		
		\hline 
	\end{tabular}
\end{table}

There is a mapping relationship between the pixel-wise orientations and the pixels in the RGB image. In other words, each pixel has its corresponding horizontal angle information and vertical angle information. Therefore, we believe that the orientations also has a mapping relationship with the image feature maps extracted by the convolutional layer, because the feature maps extracted by the CNN network is obtained by the image through the convolution operation. We scale the pixel-wise orientations to the size of the corresponding convolution feature maps and splice them into fusion features. To further verify the influence of the fusion feature on the network performance, we have done some comparison experiments. The experimental results are shown in Table \ref{table:3}. We denote the output features of the first five convolutional layers as \{conv1, conv2, conv3, conv4, conv5\}. For output feature map from each convolutional layer, we need to calculate its corresponding orientations, which is obtained by scaling the pixel-wise orientations corresponding to the original RGB. The obtained orientations is concatenated with the output feature maps, and then the fusion features are further forwarded to the next layer. As shown in Table \ref{table:3}, the prediction accuracy of the network is the highest after the orientations is concatenated with the output feature map of the third convolution layer, but its corresponding standard deviation is the lowest, which indicates that the range of steering angle predicted by this model is slightly smaller. Although the prediction accuracy of the model is not the highest after the orientations and RGB images are concatenated, the standard deviation of the steering angle predicted by the model is higher in all the comparision experiments. Considering the results in Table \ref{table:3}, we choose to concatenate the orientations with RGB image, because this model can not only maintain high prediction accuracy, but also has high standard deviation of predicted value.

\begin{table}[]
	\centering
	\caption{\textbf{Performance of models after concatenate the pixel-wise orientations with the output of different convolutional layers. } All models use SteeringLoss2 as loss function. The accuracy is measured within $5^\circ$ biases.}
	\label{table:3}
	\setlength{\tabcolsep}{3mm}{
		\begin{tabular}{ccccc}
			\hline 
			\multirow{2}{*}{} & \multicolumn{2}{c}{LiVi-Set } & \multicolumn{2}{c}{PandoraDriving} \\
			\hhline{~----}
			&Accuracy & SD & Accuracy & SD  \\
			\hline 
			RGB + VA Map + HA Map	& 67.9\% & 15.12 & 77.9\% & 25.52\\ 
			
			conv1 + VA Map + HA Map	& 68.0\% & 15.07 & 75.5\% & 23.67 \\ 
			
			conv2 + VA Map + HA Map	& 68.0\% & 14.90 &  72.3\% & 26.20 \\ 
			
			conv3 + VA Map + HA Map & 68.6\%& 14.75  & 78.1\% & 21.92 \\ 
			
			conv4 + VA Map + HA Map & 68.4\%& 15.10 & 77.2\% & 22.73\\
			
			conv5 + VA Map + HA Map & 68.5\% & 14.91  &  75.5\% & 22.14 \\ 
			\hline 
	\end{tabular}}
\end{table}

\section{Conclusions and Future Work}

In this paper, we propose a novel end-to-end steering angle control model, which takes RGB images and their corresponding pixel-wise orientations as input. In the whole process of generating the corresponding steering angle, our driving scene can be treated as the local navigation of the autonomous car, so we adopt the idea similar to the BM algorithm of robot local navigation, and propose the pixel-wise orientations of the RGB images. Then we concatenate the pixel-wise orientations with the RGB image as the network input, the orientations are similar to the sector or radial in the BM model. After adding the pixel-wise orientations, the end-to-end steering angle control model can predict the steering angle more accurately than only receiving RGB images as input. After that, In order to deal with the problem of steering angle imbalance, we propose SteeringLoss2, which is an improved version of SteeringLoss. Through experiments, we found that SteeringLoss2 has the best performance, which can not only maintain the robustness of MAE, but also avoid further amplification error of SteeringLoss, and the model based on SteeringLoss2 can estimate a wider range of steering angles. Besides, we also present our own datasets, which includes multiple driving scenarios. Finally, we conduct some comparison experiment on LiVi-Set and our own dataset, and the experiment shows that the model using our proposed methods can predict steering angle accurately.

Although our method can estimate a wider steering angle, there is still a large deviation between the estimated steering angle and the ground truth, which leads to insufficient prediction accuracy. In future work, we will feed the point clouds into the end-to-end steering angle control model. Different from two-dimensional images, the point clouds has rich depth information and geometric features, which can help the control model make better decisions.

%
%
\bibliographystyle{unsrt}
\bibliography{bibsample}

\end{document}